\documentclass[letterpaper, 10 pt, conference]{ieeeconf}  

\IEEEoverridecommandlockouts                              

\usepackage{graphics} 
\usepackage{epsfig} 
\usepackage{mathptmx} 
\usepackage{times} 
\usepackage{amsmath} 
\usepackage{amssymb}  

\usepackage[ruled,vlined]{algorithm2e}  
\usepackage{algpseudocode}  

\title{\LARGE \bf
Assembly of randomly placed parts realized by using only one robot arm with a general parallel-jaw gripper
}

\author{Jie Zhao$^{1}$, Xin Jiang$^{1}$, \emph{Member, IEEE }, Xiaoman Wang$^{1}$, Shengfan Wang$^{1}$, and  Yunhui Liu $^{1,2}$, \emph{Fellow, IEEE }
\thanks{$^{1}$Mechanical Engineering and Automation, Harbin Institute of Technology, Shenzhen 518055, China; {\tt\small zhaojie@stu.hit.edu.cn, x.jiang@ieee.org, \{16b953039,18S053234\}@stu.hit.edu.cn}}%
\thanks{$^{2}$Department of Mechanical Engineering, The Chinese University of Hong Kong, Hong Kong, China;
		{\tt\small yhliu@mae.cuhk.edu.hk}}%
}

\begin{document}

\maketitle
\thispagestyle{empty}
\pagestyle{empty}

\begin{abstract}

In industry assembly lines, parts feeding machines are widely employed as the prologue of the whole procedure. They play the role of sorting the parts randomly placed in bins to the state with specified pose.  With the help of the parts feeding machines, the subsequent assembly processes by robot arm can always start from the same condition. Thus it is expected that function of parting feeding machine and the robotic assembly can be integrated with one robot arm. This scheme can provide great flexibility and can also contribute to reduce the cost. The difficulties involved in this scheme lie in the fact that in the part feeding phase, the pose of the part after grasping may be not proper for the subsequent assembly. Sometimes it can not even guarantee a stable grasp. In this paper, we proposed a method to integrate parts feeding and assembly within one robot arm.  This proposal utilizes a specially designed gripper tip mounted on the jaws of a two-fingered gripper. With the modified gripper, in-hand manipulation of the grasped object is realized, which can ensure the control of the orientation and offset position of the grasped object. The proposal in this paper is verified by a  simulated assembly in which a robot arm completed the assembly process including parts picking from bin and a subsequent peg-in-hole assembly.      

\end{abstract}

\section{INTRODUCTION}

The demand to automatic assembly, where parts are randomly piled in bins, is increased. Unfortunately, many such processes in assembly lines need to be performed manually. Conventinally, a parts feeder cooperated with a robot is used in this situation for sorting the parts to a specified state. The solution with a parts feeder implies a specially designed mechanism and a large setting space. Since the type of dedicated parts feeders is completely determined by the type of the part used, the employment of part feeders implied high cost. In this field, a bin-picking system proposed in \cite{domae2014fast} , consisting of four robots, is used for replacing the parts feeder. In this system, the orientation alignment and assembly of the parts are performed by different robots. Although this solution can ensure the required takt time, its cost is high. It is expected that a single robot can perform both the orientation alignment and the subsequent assembly given the parts randomly placed.

\begin{figure}[tpb]
	\centering	
	\includegraphics[width=8.5cm]{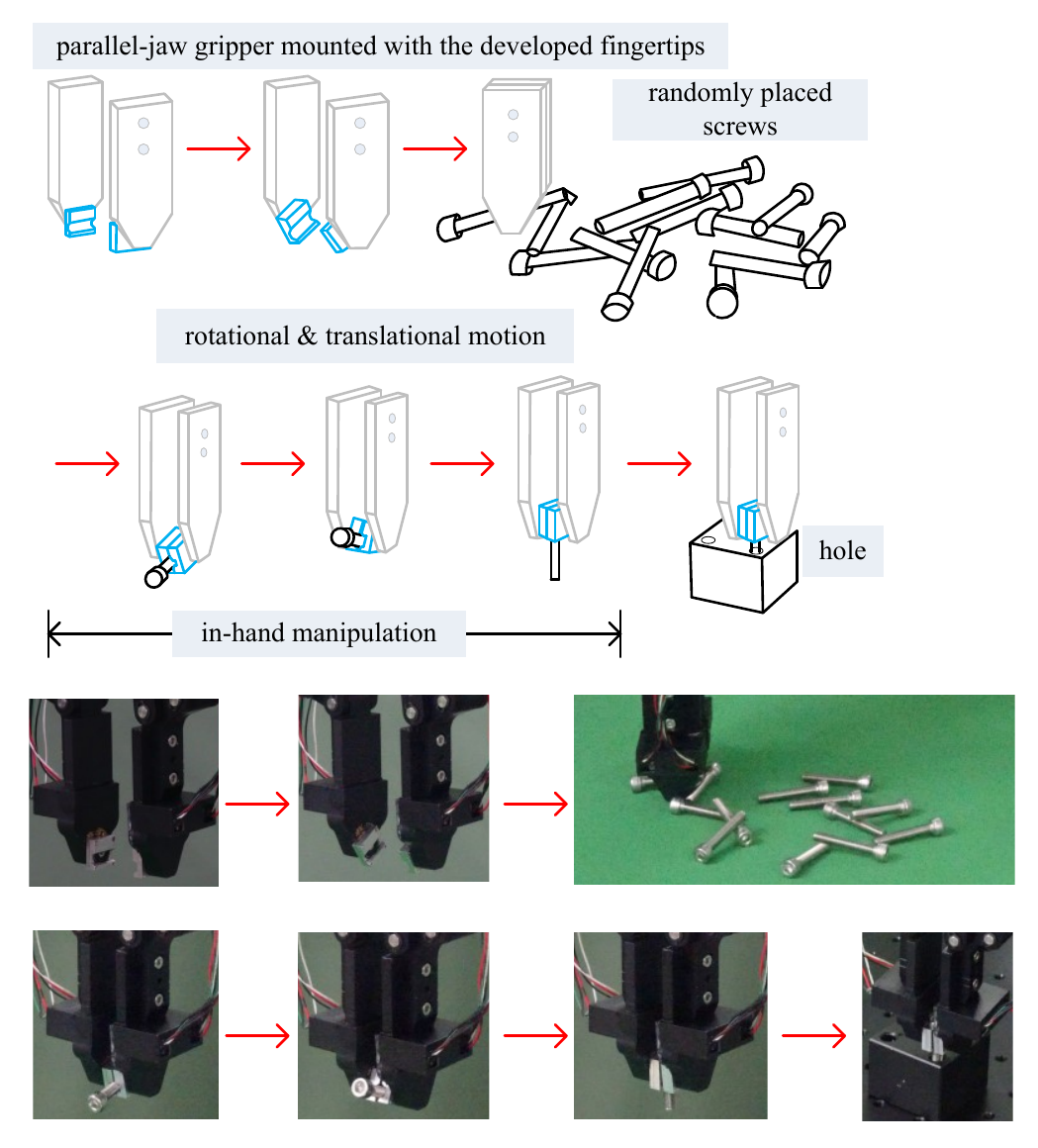}
	\caption{The whole process for integrating grasp and in-hand manipulation by only one parallel gripper mounted with the specially designed fingertips. 1. Grasp a randomly placed screw. 2. Reorientate the screw through in-hand manipulation and then insert it in a hole.
	}
	\label{figurelabe0}
\end{figure}

The goal of this research is to develop an automatic assembly system which can complete the whole assembly process including parts picking from bins and the subsequent assembly with a single robot. To realize this goal, some major challenges must be addressed: (1) grasping a randomly placed part from the cluttered bin. (2) realizing in-hand manipulation by a modified parallel jaw gripper with the developed gripper. (3) realizing force sensing during the assembly process. In our work, we mainly focus on demonstrating our approach by a simulated assembly process in which the robot firstly grasps randomly placed screws, then re-orientates it through in-hand manipulaiton, and finally inserts it in a hole, as shown in Fig.\ref{figurelabe0}. Arguably, screws assembling is the most common task on all automatic assembly lines in industry \cite{shroff2011finding}. The challenges (1) and (3) are well studied in the references \cite{shroff2011finding} and \cite{jia2018survey}. Therefore, in this paper, we primarily focus on how to realize in-hand manipulation with a simple gripper. The employment of simple grippers are reasonable in industry \cite{mason2012autonomous}.

Although with dexterous hands it can realize stable grasp, reorientation, and assembly, the control strategy involved is complex and the corresponding hardware are expensive, which are preventing dexterous hands from practical employment in industry. Recently, many studies of two or three-fingered hands have been carried out. They are not feasible for bin-picking in the condition considered in this paper. In this work, we propose to tackle the problem by installing a specially designed fingertip on a general purpose two-fingered  gripper.    

\begin{figure}[tpb]
	\centering	
	\includegraphics[width=8.5cm]{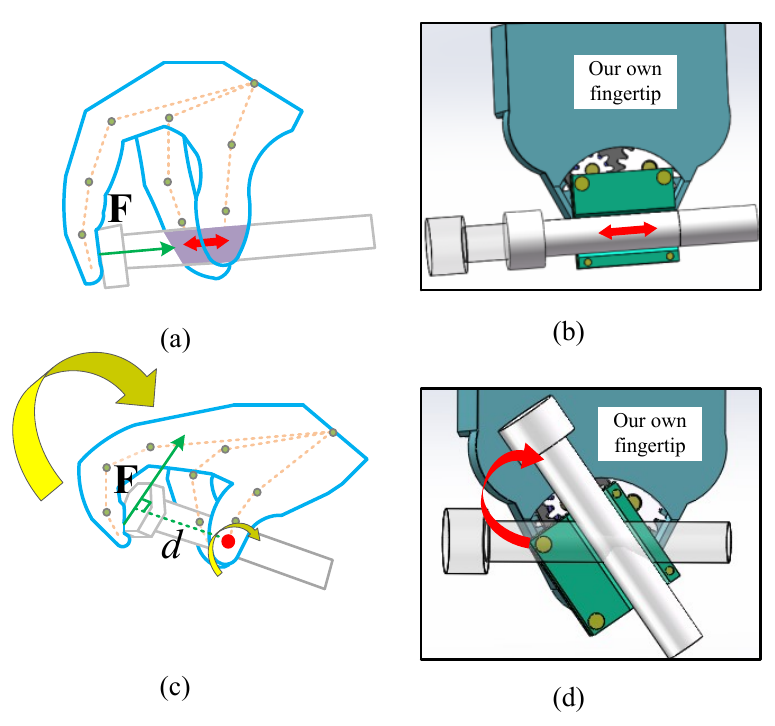}
	\caption{Comparison between human in-hand manipulation and that realized by using the proposed finger tip. (a) and (c) represent a human hand manipulation for in-hand screw. (b) Reorientate the screw through in-hand manipulation by our fingertip.
	}
	\label{figurelabe1}
\end{figure}

Our primary inspiration comes from the human hand movements in grasping, reorientating, and assembling a randomly placed part. During the above procedure, the human utilizes three fingers to complete the task through translational and rotational motion. The translational motion is realized with two of the fingers which functions as a guide for the parts. And the last finger functions to provide the pushing force, in this case. In rotating a part, two fingers function to provide pivot and the third one provide the necessary moment. The corresponding relationship between the human hand movements and that realized by the proposed gripper is shown in Fig. \ref{figurelabe1}. 

The main contributions of our paper are as follows: 

\begin{itemize}
\item A novel fingertip is proposed for in-hand manipulation of parts in assembly scenario.
\item A pose estimation method is proposed for the picked parts in in-hand manipulation scenario.
\item A strategy of utilizing the proposed gripper tip is proposed which can achieve the whole procedure from parts picking to final assembly with a single arm.
\end{itemize}

\section{RELATED WORK}

In-hand manipulation is a well studied problem in robot communities. At present, there are two main approaches for in-hand manipulation, depending on whether the approach rely on power extrinsic to the hand or not \cite{dafle2014extrinsic}. The methods that depend on either gravity, external contacts or dynamic arm motions are called extrinsic dexterity \cite{dafle2014extrinsic}. In contrast, the methods that only rely on power offered by the hand itself is called intrisinc dexterity\cite{dafle2014extrinsic}. If the contact model describing the manipulation task is relatively simple, then the manipulation task can be achieved by a planned contact sequence. In contrast, if the contact model is much more complex, then introducing a novel simple gripper for manipulaiton will be the much better alternative.

The former approach realizes in-hand manipulation through the external power to the hand. This method always includes regrasp motion during manipulaiton. However, in the industry assembly line, it should avoid collision of the parts. Nonetheless, in robot community there have been an extensive study on extrinsic dexterity. In the work of \cite{dafle2014extrinsic}, a three-fingered gripper is demonstrated to achieve in-hand manipulation through high acceleration movement of a robot arm and the gripper regrasping motion. In the above procecess, a significant adjustment of object pose could be achieved. A framework for planning the motion of a robot hand is proposed in paper \cite{shi2017dynamic}. In this way, an inertial loading on the grasped object is created to realize a sliding motion. Recently, an impressive work about extrinic dexterity  is presented in \cite{chavan2018hand}. In that paper, there is not regrasp during the whole process, and motion cones are studied in more general planar tasks, particularly in the vertical plane. By the constructed motion cones, continuous prehensile manipulaiton is realized.    

Instead of relying on external power to the hand, numerous research are interested in intrisnic dexterity. To our knowledge, a parallel-jaw gripper modified with a turntable at each fingertip is originally developed for dexterous manipulation in \cite{nagata1994manipulation}. In that work, rotating and sliding operation are achieved by controlling the turntables opposite each other. At the same time, six types of gripper with one degree of freedom at fingertips are proposed in the paper.  Tadakuma et al. \cite{tadakuma2012robotic} firstly develop a finger mechanism equipped with omnidirectional driving roller. The highlight of this finger mechanism is that the grasped object can be manipulated by roller with any arbitrary directional axes. In a recent study, Spiers et al. \cite{spiers2018variable} through analogy the human finger present a variable friction gripper fingers surfaces and realize in-hand manipulation for the grasped object.

Together these studies provide important insights into the in-hand manipulation. The former methods offer greater generality than the latter ones. However, all the extrisnic dexterity experiments are shown with the objects with flat surface. Also, the complex control, kinematics and contact model is often needed. Thus, realizing extrinsic dexterity in real world or in industry implies enormous challenges. In contrast to the extrisnic dexterity, studies on in-hand manipulation depending on power from the hand itself has a long history of application in industry, such as the fingered grippers equipped with turntable. Nevertheless, most of the research about finger grippers are not capable of in-hand manipulation, because they either have large size or not capable of rotating and translating the gripped object. 

The goal of system that is described in literature \cite{domae2014fast} is the closest one to ours. Its limitation motivates us to develop a robot system which can realize part picking and assembly with only one robot arm.

\section{METHOD}

\subsection{Realization of in-hand movement of the grasped object} 
\begin{figure}[tpb]
	\centering	
	\includegraphics[width=8.5cm]{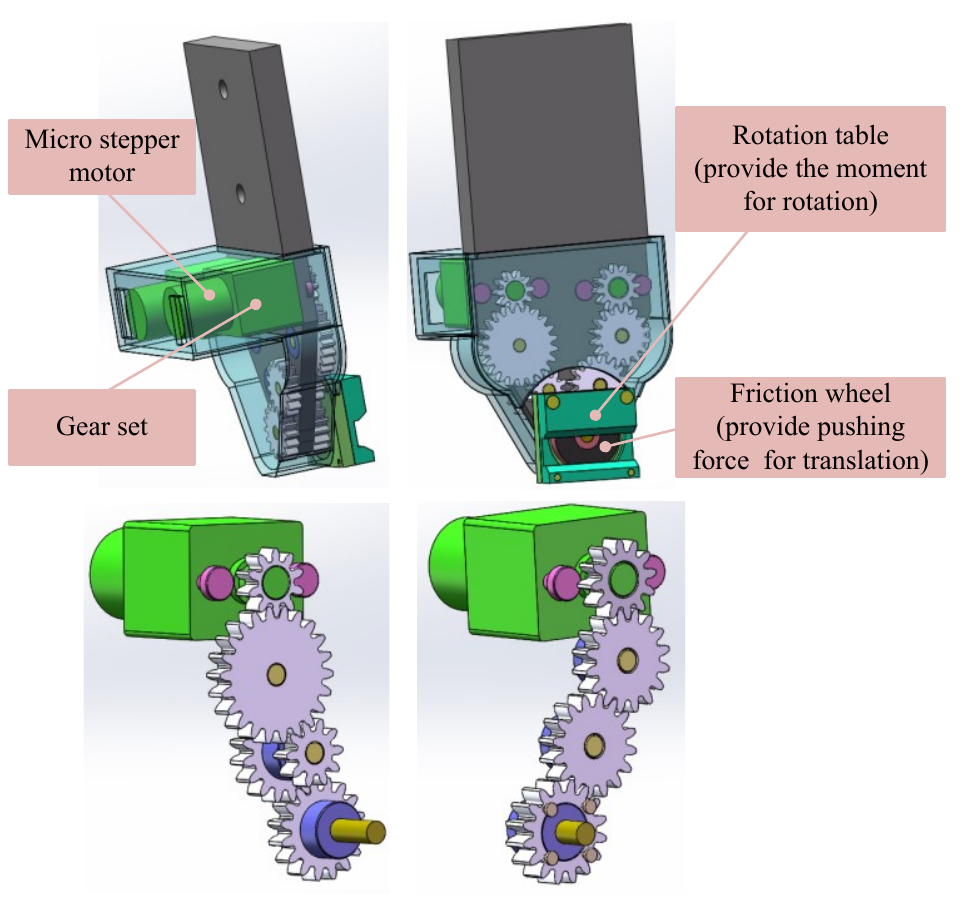}
	\caption{Side and front view, and two transmission mechanisms of gripper fingertip prototype.
	}
	\label{figurelabe2_1}
\end{figure}

By analyzing the processes of manual assembly of screws,  we find that it is generally necessary to firstly adjust the screws to a pre-assembly posture through in-hand movement, then assemble them. As shown in  Fig.\ref{figurelabe1}, we developed a gripper fingertip to emulate the actions of human hand during screw in-hand manipulation.  The whole mechanism of the gripper fingertip having two degrees of freedom as shown in Fig.\ref{figurelabe2_1}. It comprises two set of transmission mechanisms. One set is consisted of four gears and is connected to the rotational table providing the moment for a grasped object. Another set is consisted of five gears, and the last gear at the end of this set is joined with a friction wheel which provides friction force for translating a grippered object. In order to improve accuracy of in-hand manipulation, provide enough force, and reduce the complex of control, an eccentric V-shaped groove is machined in rotational table, as shown in Fig.\ref{figurelabe2_1}. 

The two transmission mechanisms make it possible for a gripper to be used in bin-picking, in-hand manipulation, and assembling. The essential point of this gripper fingertip design is the adoption of friction wheel and the groove. Based on them, a robust translational motion dramatically different from the other gripper fingertips, can be eventually acheived during intrinsic dexterity. Because of the existing eccentric, a different direction of rotation can lead to a completely diferent end-effector planning, it will be discussed in detail in the subsequent sections.

\subsection{Pre-grasp posture determination}

This section is focused on the determination of pre-grasp posture of gripper fingertips before grasping randomly placed screws. Considering the circumstances that many screws are randomly piles in bins, their postures are arbitrary. We assume that an algorithm for detecting the screws from the bin is available. By using it, we can recognize each screw and obtain its pose.

{\bfseries Calibration of the gripper tip}. In order to achieve a compact design, no absolute position sensors are embedded in the gripper fingertip. Thus when the gripper is used for the first time, it is needed to calibrate the initial rotational position of the turntable with respect to the hand. For this purpose, a Hall magnetic proximity switch sensor is introduced. The sensor is installed at the home position in our settings. When the robot moves to the home position, a flag will be set true in our programs, then the turntable rotates until the cylindrical magnet embedded in the turntable is within the sensing range of sensor.By the feedback from sensor, the rotational position of the turntable is initialized.  

\begin{figure}[tpb]
	\centering	
	\includegraphics[width=5.5cm]{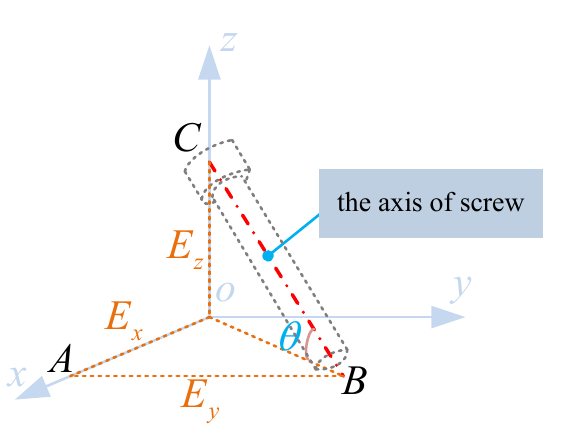}
	\caption{The coordinate system for computing the pitch angle of a randomly placed screw
	}
	\label{figurelabe3}
\end{figure}

{\bfseries Detection of the screw}. Detection of the randomly placed screws is necessary for the subsequent grasping. For this purpose, a RGB-D sensor installed facing down to the work cell is adopted. The point cloud of the crews are obtained by color based segmentation. Then, we calculate the major axis of srew by Principal Components Analysis(PCA). The Eigenvector corresponds to the largest Eigenvalue is the estimate of the central axis of srew. Next, we compute the pitch angle of screw through the three elements of eigenvector, based on the camera coordinate system as shown in Fig.\ref{figurelabe3}. The pitch angle $\theta$ can be represented as follows:
\begin{equation}
\theta = \arctan(E_z/sqrt{(E_x^2+E_y^2)})
\end{equation}
Where, $E_x$,$E_y$,$E_z$ are the projections of eigenvector on coordinate system. Then, The angular information is converted to the pulse needed for the motor controller.   	

\subsection{ In-hand manipulation }

In industry, screws are randomly piled in bins. Thus, the posture of in-hand screw is approximately the same as its initial state, and this posture is often not the ultimate desired assembly pose. Therefore, an intrisnic dexterity is very necessary in our work. The main challenges in this section are the pose estimation of in-hand screw and the strategy for intrisnic dexterity.

\begin{figure}[tpb]
	\centering	
	\includegraphics[width=5.5cm]{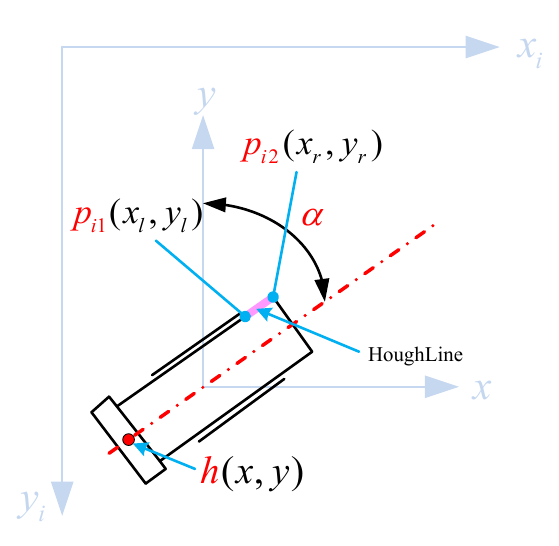}
	\caption{the bilateral rheonomic model for pose estimation for in-hand screw
	}
	\label{figurelabe3_1}
\end{figure}

{\bfseries Pose estimation for in-hand screw}. For a in-hand screw, it can be treated as a bilateral rheonomic problem \cite{mason2001mechanics}. Then the screw constrained by the finger tip would have three degrees of freedom, expressed by three variables of the coordinates $x, y$ and $\alpha$, as shown in Fig.\ref{figurelabe3_1}. In order to regulate the pose of screw with respect to the gripper tip, the robot will make the gripped screw face to another camera configured in the work cell.  Based on the image taken by the camera, the position/orientation of the screw in hand will be calculated and adjusted to the desired ones. The angle $\alpha$ between gripper and screw can be calculated by image processing. We firstly determine the region of interest (ROI) for screw detection. Through this ROI image, the notable Canny edge detection \cite{canny1986computational} is used to obtain the binary image edges. And then another notable HoughLine detection \cite{stephens1991probabilistic} is adopted in the following step for obtaining HoughLines of the extracted binary image edges. After the above image processing, the locations of two endpoints $p_{i1}(x_l,y_l)$, $p_{i2}(x_r,y_r)$ of the HoughLine segment can be obtained, where we assume that $p_{i1}$ is always on the left side of $p_{i2}$ in image. Based on the endpoints locations, the angle between screw and gripper can be represented as follows:
\begin{equation}
\alpha = \arctan(x_r-x_l/y_r-y_l)
\end{equation}
From the Fig.\ref{figurelabe3_1}, we know that the pose of in-hand screw can not be defined completely only by $\alpha$, and the location $h(x,y)$ of screw head must also be determined. By conventional image processing methods, it can not effectively solve the problem. To this end, deep learning method YOLOv3 \cite{redmon2018yolov3} is adopted for the screw head detection. This network can output the centre location of the detected bounding box. We then set the output of network as the location of screw head. The whole process of above can be found in  Algorithm 1. Based on the above two parameters, $\alpha$ and $h(x,y)$, a strategy for screw intrisnic dexterity could be designed as follows.

\begin{algorithm}  
	\caption{Pose Estimation for In-hand Screw}  
	\KwIn {an RGB image, $I$;} 
	\KwOut {rotation angle, $\alpha$; screw head location, $h(x,y)$}
	\qquad $h(x,y) \gets 0 $ \;
	\qquad $I{}'$ =   GetROI($I$)\;
	\qquad $h(x,y)$ = ScrewHeadPositionDetection($I{}'$)\; 
	\qquad ${I_e}$ = EdgeDetection($I{}'$)\;
	\qquad ${p\left[ p_{i1},p_{i2} \right]}$ =  HoughLineExtract($I_e$)\;
	\qquad $\alpha$ = $\arctan$(${x_r-x_l/y_r-y_l}$) \;
\end{algorithm}

{\bfseries Strategy for screw intrisnic dexterity}. Considering the fact that there are more than one in-hand manipulation planner for a grasped screw. In order to improve efficiency of intrisnic dexterity, an optimal planner must be determined, as shown in Algorithm 2. Let $I_m(x,y)$ denote the midpoint of fingertip in $I{}'$ as shown in Algorithm 1, it is always a constant when the setting installation is completed.

\begin{algorithm}  
	\caption{In-hand Manipulation Planner}  
	\KwIn {rotation angle, $\alpha$; screw head location, $h(x,y)$; a vector for storing two endpoints of HoughLine segment, ${p\left[ p_{i1},p_{i2} \right]}$;} 
	\KwOut {ultimate desired assemble pose, ${P_d}$}
	$I_{mx} \gets 135 $ \;
	$\pi \gets 3.14159 $ \;
	\eIf{$p.size() \neq 0$}{  
		$\Delta x \gets {x_r-x_l }$ , $\Delta y \gets {y_r-y_l} $\;
		\eIf{$\Delta y$ = $0$ }{$\alpha \gets \pi/2$;}{$\alpha \gets \arctan2$($\Delta x,\Delta y$);}
		\uIf{${h_x} > I_{mx}$ \textbf{and}  ${h_x} \neq 0$} {
			motor clockwise rotation signal, motor\_rs $\gets 0$ \;
			motor rotation angle  $\gets \alpha$ \; 
		}
		\uElseIf{${h_x} < I_{mx}$ \textbf{and}  ${h_x} \neq 0$}{
			motor clockwise rotation signal, motor\_rs $\gets 1$ \;
			motor rotation angle  $\gets \alpha$ \;
		}
		\uElseIf{${h_x} = 0$ \textbf{and}  $\Delta y > 0$}{
			motor clockwise rotation signal, motor\_rs $\gets 1$ \;
			motor rotation angle  $\gets \alpha$ \;
		}
		\Else{
			motor clockwise rotation signal, motor\_rs $\gets 0$ \;
			motor rotation angle  $\gets \alpha$ \;
		}
	}{
    motor for translation signal, motor\_ts$\gets 1$ \;
    \textbf{repeat}  Algorithm 1 \& Algorithm2\; 
    }	            
    ${P_d}$ $\gets$ Generate\_motion (motor\_rs,motor\_ts, $\alpha$) \; 
	
\end{algorithm} 

\section{EXPERIMENTS}

\begin{figure}[tpb]
	\centering	
	\includegraphics[width=8.5cm]{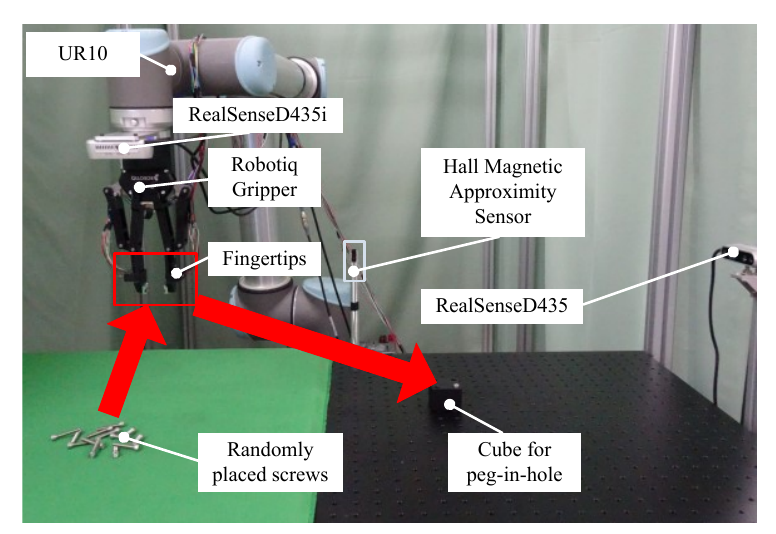}
	\caption{Experimental setup. Screws are randomly placed on a table. There are two cameras in our experiment. The camera mounted on hand is used for screw detection before grasping, another on the right of the figure is for in-hand screw pose estimation. Hall Magnetic Approximity Sensor is used for calibrating the initial pose of the proposed fingertips. 
	}
	\label{figurelabe5}
\end{figure}

In this section, we describe the implementation of our proposal. We demonstrate the manipulation capability of our prototype gripper fingertips with UR10. The verification experiments are conducted by using the setup shown in Fig. \ref{figurelabe5}. Our prototype fingertips are mounted at the end of Robotiq gripper. The system architecture is based on Robot Operating System(ROS) \cite{quigley2009ros}.

There are four subsystems in the whole experimental setup interfaced as ROS nodes:

\begin{itemize}
\item {\bfseries Robot And Gripper Controller}: Node programmed for absolute positioning of UR10 and interface with the motor controller that drives the gripper.
\item {\bfseries Fingertips Controller}: Node interfaces the stepper motor controller that drives the fingertips.
\item {\bfseries Vision System}: Node for providing 3D point cloud and RGB image to the robot and gripper controller, and providing RGB image to fingertips controller.
\item {\bfseries Learning Interface}: After offline training, this Node is programmed for pose estimation of randomly placed screws and location detection in-hand screw.
\end{itemize}

The robot follows a procedure for grasping randomly placed screws, reorientating it by in-hand manipulation and insert it in a hole. Before robot moves to the pregrasp pose, an opposite rotation of the gripper fingertips may be executed to a desired posture based on the screw pose. The screw pose in above is detected by on-hand camera. After grasping a screw, the robot will move to the next predefined pose that is denoted midpose for in-hand manipulation. Here, the side camera in our setting captures the houghline and detects screw head of the in-hand screw for computing its posture. Subsequently, a in-hand manipulation motion is performed. Following, a desired pose of screw for assembling is made to come true. Then, the robot approaches to place pose, opens the gripper, and inserts the screw in a hole. Then, robot will move to another predefined pose(we call it home\_pose) for fingertips calibration. As shown in Fig. \ref{figurelabe5}, the home\_pose is near at the postion of Hall Magnetic Approximity Sensor. From then on, a work cycle is completed. From section \uppercase\expandafter{\romannumeral3}, we know that there are more than one planner for grasping and in-hand manipulation according to the screw initial pose. We demonstrate the in-hand manipulation planner in the following.

\begin{figure}[tpb]
	\centering	
	\includegraphics[width=7cm]{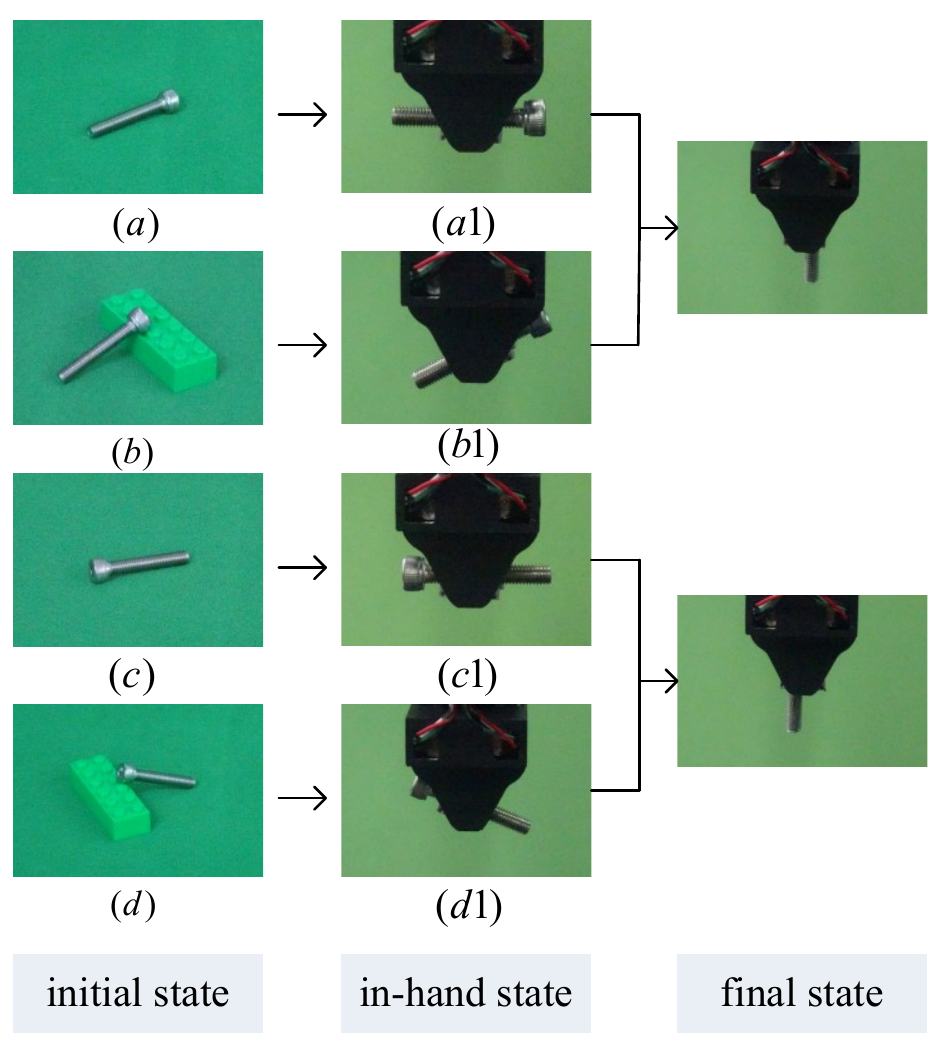}
	\caption{Four different initial postures of a randomly placed screw. (a) and (c) represent the situations where no rotation of gripper fingertips before grasping is needed. (b) and (d) represent that a rotation of our gripper fingertips is needed before grasping. In addition, (a1) and (c1) need an anticlockwise rotational motion during in-hand manipulation, in the contrary, (b1) and (d1) need a clockwise rotational motion during in-hand manipulation.
	}
	\label{figurelabe6}
\end{figure}

{\bfseries Demonstration for In-hand manipulation planner}. We mainly analyze four different initial pose of a screw. Their states can be confirmed from the Fig. \ref{figurelabe6}.

In the Fig. \ref{figurelabe6} (a) and (c), a screw is placed horizontally, it is not necessary to rotate the gripper fingertips at first. In contrast, in the situations demonstrated in  Fig. \ref{figurelabe6} (b) and (d), a screw is tilted on rectangular block. Considering this situation, a first rotation of the fingertips is needed before grasping. After grasping the screw, the robot moves to the pre-defined position in the work cell where an RBG image can be captured by RealSenseD435 for in-hand manipulation. 

\begin{figure}[pb]
	\centering	
	\includegraphics[width=8.5cm]{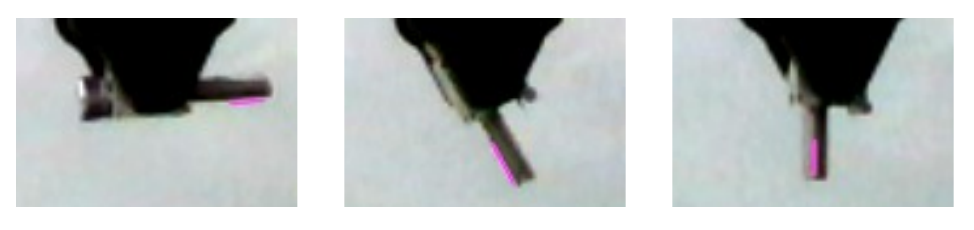}
	\caption{HoughLine detection results for three different situations of in-hand screw. The images are captured directly from realsenseD435.
	}
	\label{figurelabe6_1}
\end{figure}

For in-hand station, an image is transformed for screw head detection and computing orientation. Based on our in-hand manipulation plannar in section \uppercase\expandafter{\romannumeral3}, if no HoughLine is detected, a translational motion is needed until the edge of screw can be detected. Then, according to the detected location of screw head in image, a rotational motion is excuted for reorientation the screw from initial pose  ${p_i}$ to ultimate desired pose ${p_u}$. The detected HoughLine results for in-hand screw can be found in Fig. \ref{figurelabe6_1}

\begin{figure*}[tpb]
	\centering	
	\includegraphics[height=8.5cm]{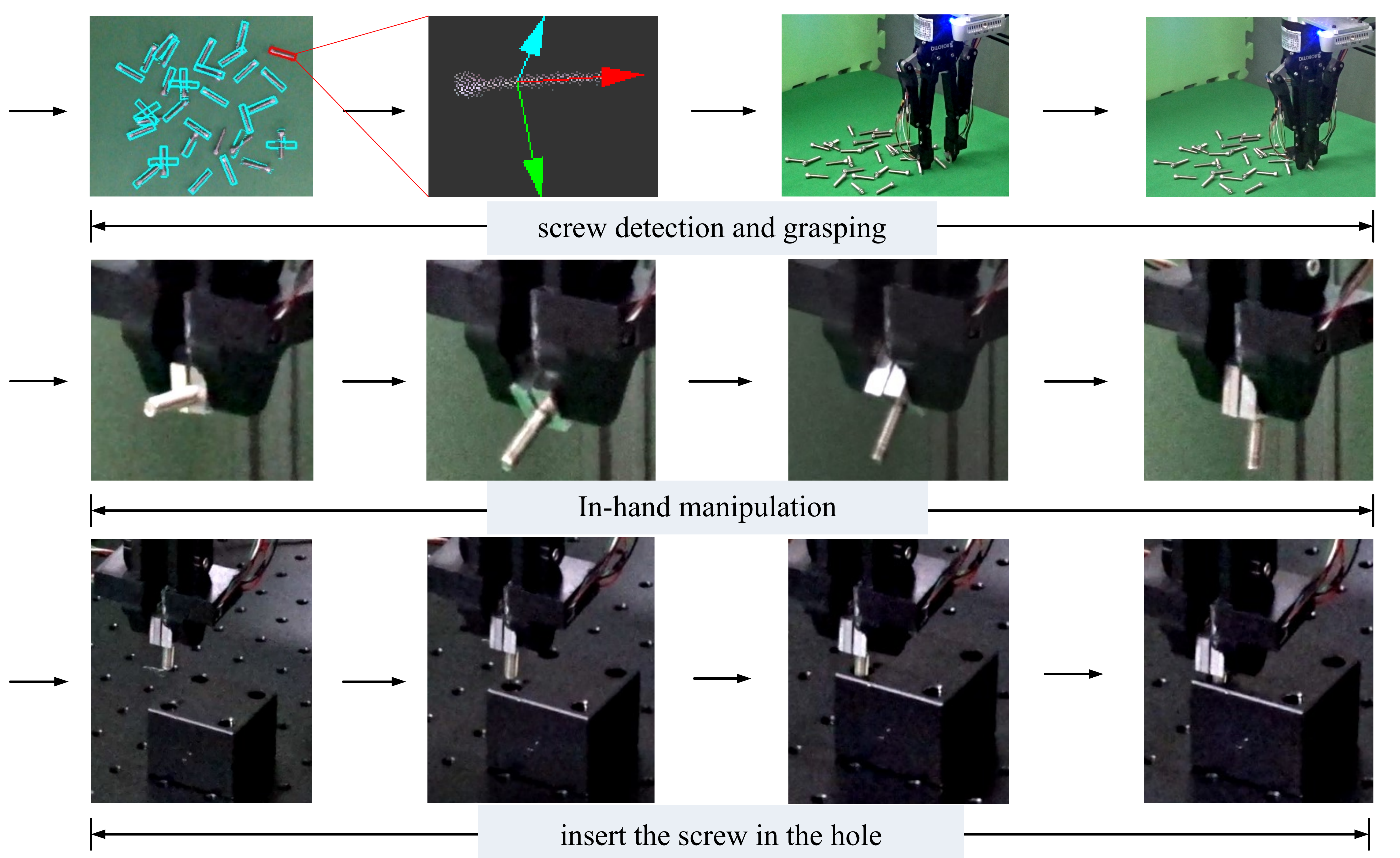}
	\caption{the processes of part picking and assembly by only one robot. The first image shows the point cloud of the screws, where the rectangulars represent the screw detection results remapping from 2D to 3D image. From the detected crews represented by the bounding box, we choose the one with the highest score as the candidate for the process.
	}
	\label{figurelabe8}
\end{figure*} 

 The experiment demonstrating a whole working cycle is conducted. The conditions of the experiments is set to simulate a real assembly in plants. Thus the robot has to complete the steps including picking screw from piled bin, position/orientation adjustment in hand and inserting the screw into the specified hole. As for the detection of multiple screws in cluttered state, we adopt the method proposed in \cite{wang2020object}. The whole process of the experiment is demonstrated in Fig. \ref{figurelabe8}.

\section{CONCLUSION}

In this work, we present an innovative architechture for integrating the grasp and in-hand manipulation procedures by only one parallel gripper. The approach is based on a specially designed fingertip which can be easily installed on the general parallel grippers. Based on this architechture, a randomly placed screw can be smoothly reorientated from its initial pose to an ultimate desired assembly pose. Experimental implementation of the designed gripper fingertips and our motion planner demonstrated the feasibility of the whole approach in assembly procedures.   

\addtolength{\textheight}{-12cm}   





\bibliographystyle{IEEEtran}
\bibliography{IEEEexample}

\end{document}